# Stable Independence in Perfect Maps

**Peter R. de Waal** and **Linda C. van der Gaag**
Institute of Information and Computing Sciences, Utrecht University
P.O. Box 80089, 3508 TB Utrecht, the Netherlands
{waal,linda}@cs.uu.nl

## Abstract

With the aid of the concept of stable independence we can construct, in an efficient way, a compact representation of a semi-graphoid independence relation. We show that this representation provides a new necessary condition for the existence of a directed perfect map for the relation. The test for this condition is based to a large extent on the transitivity property of a special form of d-separation. The complexity of the test is linear in the size of the representation. The test, moreover, brings the additional benefit that it can be used to guide the early stages of network construction.

## 1 Introduction

Probabilistic models for use in decision-support systems are typically built on the semi-graphoids axioms of independence. These axioms in fact are exploited explicitly in probabilistic graphical models, where independence is captured by topological properties, such as separation of vertices in an undirected graph or d-separation in a directed graph. Algorithms have been constructed for these graphical models that render probabilistic inference feasible by making use of the represented independences [2, 4]. A graphical representation with directed graphs has the advantage that it allows an intuitive interpretation by domain experts in terms of influences between the variables. It is not a straightforward exercise, however, to build a graphical model, either from data or from expert interviews.

Ideally a probabilistic model is represented as a graphical model in a one-to-one way, that is, independence in the one representation implies independence in the other representation. The probabilistic model then is said to be isomorphic with the graphical model, and vice versa. Pearl and Paz [3] established a set of sufficient and necessary conditions under which a probabilistic model is isomorphic with an undirected graph. The requirements on the probabilistic model for it to be undirected graph isomorphic are quite strong. More specifically, undirected graphs do not allow for the representation of induced dependencies: if a specific independence has been established to hold given some evidence, then this independence must remain valid, no matter which further evidence is obtained. In this paper we shall not consider representations of independence with undirected graphs, but focus on directed representations. Pearl [4] gave a set of necessary conditions for isomorphism of an independence relation with a directed graph. To the best of our knowledge there is no known set of sufficient conditions.

An independence relation can be fully represented by an enumeration of its statements. If the relation is isomorphic with a graphical model, then this graphical model constitutes another representation which typically is much more compact than an enumeration of its statements. Studený introduced an alternative, generally applicable representation by means of a partial order on the independence relation [5]. All statements in the relation can be derived from the dominant statements in this order, and the set of dominant statements thus fully captures the relation. In [7] we extended this idea by introducing an additional partial order on a subset of the independence relation. This order exploits the property that some independence statements are stable, in the sense that they exclude further induced dependences. The two partial orders combined allow for a compact representation of any independence relation in general.

In this paper we compare graphical representations and representations with dominant statements and study how they are related. We show that dominance of independence statements translates to properties of a graphical model. We further show that stable independence can be translated into a special form of d-separation, which has stronger properties than ordinary d-separation. These properties lead to the formulation of a new necessary condition for an independence relation to be isomorphic with a directed graphical model. This new condition is not implied by Pearl's set of conditions. Moreover, the complexity of the test for this condition is linear in the size of the representation with dominant statements.

The paper is organised as follows. In Section 2 we briefly review the representation of a semi-graphoid independence relation by its set of dominant statements. In Section 3 we introduce the concept of strong d-separation in directed graphs, and in Section 4 we derive some interesting properties for it. In Section 5 we address the relation between strong d-separation in a graphical model and the representation of a semi-graphoid independence relation by dominant statements. In Section 6 we discuss the implications of our results with respect to network construction. In Section 7 we wrap up with conclusions and recommendations.

## 2 Stability revisited

We consider a finite index set $V = \{1,\ldots,v\}$, $v \geq 1$, where each index denotes a statistical variable. The set of ordered triplets $\langle X, Y | Z \rangle$, $X, Y \neq \varnothing$, of pairwise disjoint subsets of $V$ is denoted by $\mathcal{T}(V)$. For simplicity of notation we will often write $XY$ to denote the union $X \cup Y$ and even $Xy$ to denote the union $X \cup \{y\}$, for $X, Y \subset V$, $y \in V$. We shall use the notation $\mathcal{I}\langle X, Y | Z \rangle$ to indicate $\langle X, Y | Z \rangle \in \mathcal{I}$, for any ternary relation $\mathcal{I} \subset \mathcal{T}(V)$.

A ternary relation $\mathcal{I} \subset \mathcal{T}(V)$ is a *semi-graphoid independence relation*, or semi-graphoid for short, if it satisfies the following four axioms:

**A1:** $\mathcal{I}\langle X, Y | Z \rangle \to \mathcal{I}\langle Y, X | Z \rangle$;
**A2:** $\mathcal{I}\langle X, YW | Z \rangle \to \mathcal{I}\langle X, Y | Z \rangle \wedge \mathcal{I}\langle X, W | Z \rangle$;
**A3:** $\mathcal{I}\langle X, YW | Z \rangle \to \mathcal{I}\langle X, Y | WZ \rangle$;
**A4:** $\mathcal{I}\langle X, Y | Z \rangle \wedge \mathcal{I}\langle X, W | YZ \rangle \to \mathcal{I}\langle X, YW | Z \rangle$;

for all sets of variables $X$, $Y$, $Z$, $W \subset V$. A statement $\mathcal{I}\langle X, Y | Z \rangle$ then is taken to mean that $X$ and $Y$ are independent given $Z$ in $\mathcal{I}$. The four axioms are termed the *symmetry* (A1), *decomposition* (A2), *weak union* (A3), and *contraction axiom* (A4), respectively. Together they are referred to as the *semi-graphoid axioms*. The semi-graphoid axioms are logically independent and they are satisfied by any ternary relation $\mathcal{I}$ that is defined by probabilistic conditional independence [1, 4].

Given a set of independence statements a complete independence relation can be constructed by iteratively applying the semi-graphoid axioms. Now, consider a set of independence statements that have been established from expert interviews and/or data analysis. To build a graphical model for the independence relation that is defined by these statements, the following steps must be taken:

1. Determine the entire independence relation, i.e. the set of independence statements that includes the given set and is closed under the semi-graphoid axioms.

2. Find a graphical model that captures the relation that was obtained in Step 1 as closely as possible.

The ideal graphical representation would be one in which the topology of the graph provides for the definition of a ternary relation on its vertices, that is equivalent to the independence relation. Such a graph is called isomorphic with the relation. In practice, unfortunately, such a graphical representation often does not exist. We shall discuss isomorphism in more detail in Section 3.

The first step in the procedure outlined above amounts to computing the so-called semi-graphoid closure of a set of independence statements, which is defined as follows.

**Definition 2.1 (Semi-graphoid closure)** *Let $\mathcal{I} \subset \mathcal{T}(V)$ be a ternary relation on $V$. Then, $\mathrm{sem}(\mathcal{I})$ is the closure of $\mathcal{I}$ under the semi-graphoid axioms, that is,*

$$\mathrm{sem}(\mathcal{I}) = \bigcap_{\substack{\mathcal{I} \subseteq \mathcal{M} \subseteq \mathcal{T}(V) \\ \mathcal{M} \text{ is a semi-graphoid}}} \mathcal{M}.$$

Studený defined a concept of dominance for triplets [5], that allows a representation of $\mathrm{sem}(\mathcal{I})$ that is more compact than enumeration of its members.

**Definition 2.2 (O-dominance)** *Let $\langle T, U | W \rangle$, $\langle X, Y | Z \rangle \in \mathcal{T}(V)$. We say that $\langle X, Y | Z \rangle$ o-dominates $\langle T, U | W \rangle$, denoted $\langle T, U | W \rangle \prec \langle X, Y | Z \rangle$, if $T \subseteq X$, $U \subseteq Y$, and $Z \subseteq W \subseteq XYZ$. Now, let $\mathcal{I} \subset \mathcal{T}(V)$ be a ternary relation on $V$. A triplet in $\mathcal{I}$ that is not o-dominated by any other triplet in $\mathcal{I}$ is termed* maximally o-dominant *in $\mathcal{I}$.*

From the conditions of o-dominance we observe that a statement $u \in \mathcal{T}(V)$ is o-dominated by a statement $w \in \mathcal{T}(V)$, if $u$ can be derived from $w$ by applying the decomposition and weak union axioms.

Studený showed that the semi-graphoid closure of a given set of independence statements $\mathcal{I}$ can be represented by the set $\mathcal{D}_\mathcal{I}$ of maximally o-dominant triplets:

$$\mathrm{sem}(\mathcal{I}) = \{u \,|\, \exists_{w \in \mathcal{D}_\mathcal{I}} : u \prec w\}$$

Moreover, this set of maximally o-dominant triplets can be computed much more efficiently than exhaustive application of the semi-graphoid axioms on the given set $\mathcal{I}$ [5].

In a semi-graphoid independence relation in general we can often distinguish statements for which induced dependences are possible and statements for which they are not. In [7] we introduced the concept of stable independence to capture this difference: an independence statement is stable if it does not allow any further induced dependences.

**Definition 2.3 (Stability)** *Let $\mathcal{I} \subset \mathcal{T}(V)$ be a semi-graphoid independence relation on $V$. Then,*

- *an independence statement $\mathcal{I}\langle X, Y | Z \rangle$ is* stable *in $\mathcal{I}$ if $\mathcal{I}\langle X, Y | Z' \rangle$ for all sets $Z'$ with $Z \subset Z'$; if $XYZ = V$, then $\mathcal{I}\langle X, Y | Z \rangle$ is called* trivially stable*;*
- *an independence statement $\mathcal{I}\langle X, Y | Z \rangle$ is called* unstable *in $\mathcal{I}$ if it is not stable in $\mathcal{I}$.*

*The set of all triplets that are stable in $\mathcal{I}$ is called the* stable part *of $\mathcal{I}$, and will be denoted by $\mathcal{S}_\mathcal{I}$; the set of all unstable triplets is called the* unstable part *of $\mathcal{I}$, denoted $\mathcal{U}_\mathcal{I}$.*

For simplicity of notation we shall often write $\mathcal{S}_\mathcal{I}\langle A, B | C \rangle$ as a shorthand for $\langle A, B | C \rangle \in \mathcal{S}_\mathcal{I}$. We showed that the stable part of an independence relation satisfies the semi-graphoid axioms and hence, is a semi-graphoid independence relation itself. In addition it satisfies the composition/decomposition axiom (A2S) and the strong union axiom (A5):

**A2S:** $\mathcal{S}_\mathcal{I}\langle X, YW | Z \rangle \leftrightarrow \mathcal{S}_\mathcal{I}\langle X, Y | Z \rangle \wedge \mathcal{S}_\mathcal{I}\langle X, W | Z \rangle$;
**A5:** $\mathcal{S}_\mathcal{I}\langle X, Y | Z \rangle \rightarrow \mathcal{S}_\mathcal{I}\langle X, Y | ZW \rangle$;

for all sets of variables $X$, $Y$, $Z$, $W \subset V$. The composition/decomposition axiom (A2S) is actually the bi-implication of decomposition (A2) for stable independence statements.

Analogous to the concepts of semi-graphoid closure and o-dominance we defined the stable semi-graphoid closure and stable dominance.

**Definition 2.4 (Stable closure)** *Let $\mathcal{I} \subset \mathcal{T}(V)$ be a ternary relation on $V$. Then, $\mathrm{stab}(\mathcal{I})$ is the closure of $\mathcal{I}$ under the stable semi-graphoid axioms, that is,*

$$\mathrm{stab}(\mathcal{I}) = \bigcap_{\substack{\mathcal{I} \subseteq \mathcal{M} \subseteq \mathcal{T}(V) \\ \mathcal{M} \text{ is a stable semi-graphoid}}} \mathcal{M}.$$

**Definition 2.5 (S-dominance)** *Let $\langle T, U | W \rangle, \langle X, Y | Z \rangle \in \mathcal{T}(V)$. We say that $\langle X, Y | Z \rangle$ s-dominates $\langle T, U | W \rangle$, denoted $\langle T, U | W \rangle \lll \langle X, Y | Z \rangle$, if $T \subseteq X$, $U \subseteq Y$, and $Z \subseteq W$. Now, let $\mathcal{I} \subset \mathcal{T}(V)$ be a ternary relation on $V$. A triplet that is not s-dominated by any other triplet in $\mathcal{I}$ is termed* maximally s-dominant *in $\mathcal{I}$.*

From the conditions of s-dominance it is readily seen that a statement $u$ is s-dominated by a statement $w$ if $u$ can be derived from $w$ by applying the decomposition and strong union axioms. The stable semi-graphoid closure of a given set of independence statements $\mathcal{I}$ can now be represented by the set $D_\mathcal{I}^S$ of maximally s-dominant triplets:

$$\mathrm{stab}(\mathcal{I}) = \{u \mid \exists_{w \in \mathcal{D}_\mathcal{I}^S} : u \lll w\}$$

In [7] we gave an efficient algorithm to establish $D_\mathcal{I}^S$.

The representation of $\mathrm{sem}(\mathcal{I})$ by o-dominant triplets and that of $\mathrm{stab}(\mathcal{I})$ by s-dominant triplets can be combined.

We assume that $\mathcal{I}$ is partitioned into a set $\mathcal{I}^S$ of stable independence statements and a set $\mathcal{I}^U$ of independence statements for which stability has not been established. The set of all independence statements that can be generated from $\mathcal{I}$ by the semi-graphoid axioms can now be represented by a set $\mathcal{D}_\mathcal{I}^S$ of maximally s-dominant triplets and a set $\mathcal{D}_\mathcal{I}^U$ of maximally o-dominant triplets, such that

$$\mathrm{sem}(\mathcal{I}) = \mathrm{sem}\left(\mathcal{I}^U \cup \mathrm{stab}\left(\mathcal{I}^S\right)\right) =$$
$$= \left\{u \mid \left(\exists_{v \in \mathcal{D}_\mathcal{I}^U} : u \prec v\right) \vee \left(\exists_{v \in \mathcal{D}_\mathcal{I}^S} : u \lll v\right)\right\}$$

We presented an algorithm for the computation of $\mathcal{D}_\mathcal{I}^U$ and $\mathcal{D}_\mathcal{I}^S$ in [7]. We further showed that the representation of an independence relation by $\mathcal{D}_\mathcal{I}^U$ and $\mathcal{D}_\mathcal{I}^S$ is more compact than the representation that uses maximally o-dominant triplets only.

## 3 Directed acyclic graphs and strong d-separation

In this section we discuss the relationship between directed acyclic graphs and the representation of stability. More specifically we shall formulate a notion of separation that provides a graphical equivalent of stable independence. Before introducing this new notion we first review the standard concepts of blocking and d-separation in directed graphs.

We consider a directed acyclic graph (DAG) $G = (V, A)$, with $V$ the set of variables and $A$ the set of arcs. Let $Z$ be a subset of $V$. We say that a chain $s$ is *blocked* by $Z$ in $G$, if $s$ contains three consecutive variables $V_1$, $V_2$, and $V_3$ for which one of the following conditions holds:

- $s$ has arcs $V_1 \leftarrow V_2$ and $V_2 \rightarrow V_3$, and $V_2 \in Z$;
- $s$ has arcs $V_1 \rightarrow V_2$ and $V_2 \rightarrow V_3$, and $V_2 \in Z$;
- $s$ has arcs $V_1 \rightarrow V_2$ and $V_2 \leftarrow V_3$, and $\sigma^*(V_2) \cap Z = \varnothing$, where $\sigma^*(V_2)$ includes $V_2$ and all its descendants.

While the concept of blocking is defined for a single chain, the d-separation criterion applies to the set of all chains in $G$. Let $X, Y, Z \subset V$, $X, Y \neq \varnothing$, be mutually disjoint sets of variables. The set $Z$ now is said to *d-separate* the sets $X$ and $Y$, denoted $\langle X, Y | Z \rangle_G^d$, if for every chain $s$ between any variable from $X$ and any variable from $Y$, we have that $s$ is blocked by $Z$ in $G$.

Based on the d-separation criterion the notion of an independence model is defined: the *graphical independence model* $M_G$ of a DAG $G$ is the set of statements $\langle X, Y | Z \rangle$ such that $\langle X, Y | Z \rangle \in M_G$ if and only if $\langle X, Y | Z \rangle_G^d$, for all mutually disjoint sets of variables $X, Y, Z \subset V$, $X, Y \neq \varnothing$.

The various statements of a graphical independence model are captured by the topology of the graph. For a semi-graphoid independence relation in general the independence statements are captured by a set of generating statements from which the entire relation can be constructed by application of the semi-graphoid axioms. While every graphical independence model satisfies the semi-graphoid axioms, the reverse property does not hold, that is, not every semi-graphoid independence relation can be fully represented by a graphical model. We now say that a semi-graphoid independence relation $\mathcal{I}$ is *DAG-isomorphic*, if there exists an acyclic digraph $G$, such that

$$\mathcal{I}\langle X, Y | Z \rangle \Leftrightarrow \langle X, Y | Z \rangle_G^d,$$

for any disjoint $X$, $Y$, $Z \subset V$ with $X, Y \neq \emptyset$. Such a graph $G$ then is called a *directed perfect map* (or *directed P-map* for short) of $\mathcal{I}$. There exists a set of necessary conditions for a semi-graphoid independence relation to be DAG-isomorphic. We shall discuss these conditions in more detail in Section 5.

We now distinguish between two ways of blocking a chain, which are related to the graphical representations of unstable and stable independence.

**Definition 3.1 (Blocking by presence of information)**
*Let $G = (V, A)$ be an acyclic digraph. Let $s$ be a chain in $G$ and let $Z \subset V$. Then, $s$ is blocked by $Z$ in $G$ by* presence of information *if $s$ contains three consecutive variables $V_1$, $V_2$, $V_3$, for which one of the following conditions holds:*

- *$s$ has arcs $V_1 \leftarrow V_2$ and $V_2 \rightarrow V_3$, and $V_2 \in Z$;*
- *$s$ has arcs $V_1 \rightarrow V_2$ and $V_2 \rightarrow V_3$, and $V_2 \in Z$.*

*The chain $s$ is blocked by* absence of information *if $s$ is blocked by $Z$ in $G$ and $s$ is not blocked by $Z$ in $G$ by presence of information.*

Building on the two different ways of blocking a chain, we distinguish between strong and weak d-separation.

**Definition 3.2 (Strong d-separation)** *Let $G = (V, A)$ be an acyclic digraph, and let $X$, $Y$, $Z \subset V$, $X, Y \neq \emptyset$, be mutually disjoint sets of variables. The set $Z$ is said to* strongly d-separate *$X$ and $Y$ in $G$, denoted $\langle X, Y | Z \rangle_G^{S_d}$, if every chain between any variable from $X$ and any variable from $Y$ is blocked in $G$ by $Z$ by presence of information. The set $Z$ is said to* weakly d-separate *$X$ and $Y$ in $G$, denoted $\langle X, Y | Z \rangle_G^{W_d}$, if $Z$ d-separates $X$ and $Y$ without strongly d-separating them.*

From the above definition it is readily seen that strong d-separation implies ordinary d-separation, and that ordinary d-separation implies either strong or weak d-separation. It is also immediate that if two sets of variables $X$ and $Y$ are strongly d-separated by some set $Z$, then $X$ and $Y$ will remain d-separated if $Z$ is replaced by any superset $Z' \supset Z$. Strong d-separation therefore matches the strong union axiom.

We conclude this section with the definition of a graphical strong independence model.

**Definition 3.3 (Graphical strong independence model)**
*The graphical strong independence model $M_G^S$ of an acyclic digraph $G$ is the set of statements $\langle X, Y | Z \rangle$ such that $\langle X, Y | Z \rangle \in M_G^S$ if and only if $\langle X, Y | Z \rangle_G^{S_d}$, for all mutually disjoint sets of variables $X$, $Y$, $Z \subset V$, $X, Y \neq \emptyset$.*

From Definition 3.3 it is clear that if a semi-graphoid independence relation $\mathcal{I}$ is DAG-isomorphic with a given graph $G$, then the stable part of $\mathcal{I}$ is equal to the strong independence model $M_G^S$ of $G$.

## 4 Properties of strong d-separation

In this section we shall investigate the graphical properties of the concept of strong d-separation. We shall establish, more specifically, that strong d-separation in directed graphs satisfies the properties of separation in undirected graphs. Although this is an interesting result in itself, the importance of the strong d-separation properties lies in their translation into properties for the dominant triplets of the strong independence model of a digraph, as will be discussed in Section 5.

The first property of interest relates the concept of strong d-separation to ordinary graph-theoretical separation. From this property we have that strong d-separation in a directed graph behaves like separation in an undirected graph.

**Lemma 4.1 (Separation)** *Let $G = (V, A)$ be an acyclic digraph. If for some mutually disjoint sets $X$, $Y$, $Z \subset V$, with $X, Y \neq \emptyset$, the sets $X$ and $Y$ are strongly d-separated by $Z$, then $Z$ also separates $X$ and $Y$, in the sense that every chain between any variable in $X$ and any variable in $Y$ includes at least one variable from $Z$.*

*Proof.* It is clear that a chain between $X$ and $Y$ can only be blocked by $Z$ in $G$ by presence of information, if it has at least one variable in $Z$. □

The second property of strong d-separation is transitivity.

**Theorem 4.2 (Transitivity)** *Let $G = (V, A)$ be an acyclic digraph. Strong d-separation in $G$ satisfies the transitivity property, that is, if for three mutually disjoint sets $X$, $Y$, $Z \subset V$, $X, Y \neq \emptyset$, $X$ and $Y$ are strongly d-separated by $Z$, then any variable $\gamma \notin XYZ$ is also strongly d-separated by $Z$, from either $X$ or $Y$ or both.*

*Proof.* We prove the contrapositive form of the transitivity property, that is,

$$\exists_{\gamma \notin XYZ} \left[ \neg \langle \gamma, Y | Z \rangle_G^{S_d} \wedge \neg \langle X, \gamma | Z \rangle_G^{S_d} \right] \rightarrow \neg \langle X, Y | Z \rangle_G^{S_d}$$

Assume that for some $\gamma \notin XYZ$ we have $\neg \langle X, \gamma | Z \rangle_G^{S_d}$ and $\neg \langle \gamma, Y | Z \rangle_G^{S_d}$. From the definition of strong d-separation we can conclude that there exists a chain between $X$ and $\gamma$ that is not blocked by $Z$ by presence of information. This chain is either not blocked by $Z$, or it is blocked by $Z$ by absence of information. A similar chain must exist between $\gamma$ and $Y$. By concatenating these two chains we have a chain from $X$ to $Y$, that is not blocked by $Z$ by presence of information. We conclude that $X$ and $Y$ cannot be strongly separated by $Z$. □

As an alternative to the contrapositive proof above Theorem 4.2 also allows a direct proof, which is much more elaborate. We briefly review it here, as it provides insight into the structure of a graphical strong independence model. The proof refers to Figure 1. Without loss of generality we assume that the digraph $G$ is connected. The nodes in the figure represents disjoint subsets of $V$. An edge between two subsets indicates that in the original digraph there exists a chain between these two subsets that has no variables that are not included in the union of these two subsets. From Figure 1 we read, for instance, that there exists a chain in $G$ from $C$ to $X$ that has no variables from $V \setminus CX$, and that any chain between $C$ and $Y$ must have at least one variable from $V \setminus CY$.

Now assume $\langle X, Y | Z \rangle_G^{S_d}$, and let $\gamma \in V \setminus XYZ$. The separation property of strong d-separation (Lemma 4.1) implies that the sets $A$ and $B$ in Figure 1 are empty. For the remaining possible locations for $\gamma$ we can thus distinguish between three cases:

a. There exists a chain between $\gamma$ and $X$, that has no vertices in $Z$, i.e. $\gamma$ is in $C$ or in $E$;

b. There exists a chain between $\gamma$ and $Y$, that has no vertices in $Z$, i.e. $\gamma$ is in $D$ or in $G$;

c. All chains between $\gamma$ and $XY$ must pass through $Z$, i.e. $\gamma$ is in $F$.

For the three cases above the following properties are satisfied:

a. $\langle \gamma, Y | Z \rangle_G^{S_d}$;

b. $\langle X, \gamma | Z \rangle_G^{S_d}$;

c. $\langle \gamma, Y | Z \rangle_G^{S_d}$ or $\langle X, \gamma | Z \rangle_G^{S_d}$ or both.

The proof for case a proceeds as follows. Let $s$ be a chain between $\gamma$ and $Y$. If $\gamma \in E$, then $s$ must include at least one variable from $X$. We conclude that $s$ must be blocked by $Z$ by presence of information, since $\langle X, Y | Z \rangle_G^{S_d}$. If $\gamma \in C$, then there must exist a chain $t$ between $\gamma$ and $X$. Concatenating $s$ and $t$ gives a new chain $s'$ between $X$ and $Y$ which must be blocked by $Z$ by presence of information. Since $t$ does not include any variable from $Z$, $s$ must be blocked

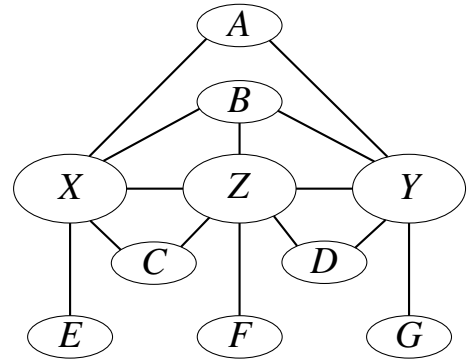

Figure 1: Visualisation of the separation and transitivity properties

by $Z$ by presence of information. The proof for case b proceeds in a similar manner. For case c we need to distinguish between three subcases:

c1 There exists a chain between $\gamma$ and $X$ that is not blocked by $Z$ in $G$ by presence of information.

c2 There exists a chain between $\gamma$ and $Y$ that is not blocked by $Z$ in $G$ by presence of information.

c3 All chains between $\gamma$ and $XY$ are blocked by $Z$ in $G$ by presence of information.

We first prove that cases c1 and c2 are mutually exclusive. To this end we assume the contrary, i.e. that there exist two chains, namely $s_X$ between $\gamma$ and $X$ and $s_Y$ between $\gamma$ and $Y$ that are both not blocked by $Z$ in $G$ by presence of information. By concatenating $s_X$ and $s_Y$, we find a chain between $X$ and $Y$ that is not blocked by $Z$ by presence of information, which contradicts the assumption $\langle X, Y | Z \rangle_G^{S_d}$.

Now assume that the conditions of case c1 hold. Since c1 excludes c2, it implies that all chains between $\gamma$ and $Y$ are blocked by $Z$ by presence of information, i.e. $\langle \gamma, Y | Z \rangle_G^{S_d}$. If the conditions of c2 hold, then we get in an analogous manner $\langle X, \gamma | Z \rangle_G^{S_d}$. If the conditions of c3 hold, then we find both $\langle \gamma, Y | Z \rangle_G^{S_d}$ and $\langle X, \gamma | Z \rangle_G^{S_d}$. This concludes the direct proof.

The usefulness of the direct proof of Theorem 4.2 will become apparent after Proposition 4.4, when we combine the transitivity property with the following property of composition for strong d-separation.

**Lemma 4.3 (Composition)** *Let $G = (V, A)$ be an acyclic digraph. Strong d-separation in $G$ satisfies the composition property, that is, for every three mutually disjoint sets $X, Y, Z \subset V$, $X, Y \neq \emptyset$, we have that*

$$\langle X, Y | Z \rangle_G^{S_d} \wedge \langle X, W | Z \rangle_G^{S_d} \Rightarrow \langle X, YW | Z \rangle_G^{S_d}$$

*Proof.* The composition property follows directly from the definition of strong d-separation. □

The transitivity property and the composition property can be combined into the following proposition.

**Proposition 4.4** *Let $G = (V, A)$ be an acyclic digraph. Strong d-separation in $G$ satisfies the property that for every three mutually disjoint sets $X, Y, Z \subset V$, with $X, Y \neq \emptyset$, we have that*

$$\langle X, Y | Z \rangle_G^{S_d} \Rightarrow \langle X, Y\gamma | Z \rangle_G^{S_d} \vee \langle X\gamma, Y | Z \rangle_G^{S_d},$$

*for each $\gamma \in V \backslash XYZ$.*

*Proof.* The proof is immediate from the combination of Theorem 4.2 and Lemma 4.3. □

In Section 5 the property from Proposition 4.4 will be translated into a test for the existence of a directed P-map for a semi-graphoid independence relation. Referring again to Figure 1 Proposition 4.4 implies that if $X$ and $Y$ are strongly d-separated in $G$ by $Z$, then $X$ can be extended to $X' = XCE$, and $Y$ can be extended to $Y' = YDG$ without destroying their strongly d-separation by $Z$. The proposition further states that all the variables from $F$ can be added to $X$, to $Y$, or to both. These variables in $F$ cannot all be added to the same sets though, as this is determined by the directions of the arcs in the chains that connect the variables in $F$ via $Z$ to $X$ or $Y$.

We conclude this section by showing that strong d-separation satisfies the intersection property.

**Theorem 4.5 (Intersection)** *Let $G = (V, A)$ be an acyclic digraph. Strong d-separation satisfies the intersection property, i.e. for any mutually disjoint sets $X$, $Y$, $W$, $Z \subset V$ with $X, Y, W \neq \emptyset$, we have*

$$\langle X, Y | ZW \rangle_G^{S_d} \wedge \langle X, W | ZY \rangle_G^{S_d} \Rightarrow \langle X, YW | Z \rangle_G^{S_d}.$$

*Proof.* Let $s$ be a chain between $X$ and $YW$. We assume that $s$ is the ordered sequence of variables $s = (x_1, \ldots, x_j, z_1, \ldots, z_k, \gamma_1, \ldots, \gamma_l)$, for some $j, l \geq 1$, $k \geq 0$, with $x_1, \ldots, x_j \in X$, $z_1, \ldots, z_k \in Z$, and $\gamma_1, \ldots, \gamma_l \in YW$. We assume, without loss of generality, that $\gamma_1 \in Y$. The subchain $s' = (x_1, \ldots, x_j, z_1, \ldots, z_k, \gamma_1)$ is a chain between $X$ and $Y$. Since $X$ and $Y$ are strongly d-separated by $ZW$, and $s'$ does not include any variables from $W$, this chain $s'$ is blocked by $Z$ in $G$ by presence of information. We conclude that $s$ also is blocked by $Z$ in $G$ by presence of information, and hence $\langle X, YW | Z \rangle_G^{S_d}$. □

## 5 Maximally dominant triplets in perfect maps

Having studied the properties of strong d-separation in the previous section we now address the relation between strong d-separation and maximally dominant triplets. The main result for a representation with o-dominant triplets states that if a semi-graphoid independence relation is DAG-isomorphic, it must include at least one o-dominant statement that involves all the variables in $V$.

We start by showing that a DAG-isomorphic independence relation contains at least one so-called saturated independence statement.

**Definition 5.1 (Saturated independence statement)** *An independence statement $\langle X, Y | Z \rangle$ over the variable set $V$ is called* saturated *if $XYZ = V$.*

**Lemma 5.2** *Let $\mathcal{I}$ be a DAG-isomorphic independence relation. Then, there exists an independence statement in $\mathcal{I}$ that is saturated.*

*Proof.* The proof is along the lines of [6, Theorem 1]. Let $G = (V, A)$ be a directed P-map of the independence relation $\mathcal{I}$. Since $G$ is a DAG, there exists a partial order $\triangleleft$ on $V$, such that, for each pair $v_1, v_2 \in V$, if there is an arc from $v_1$ to $v_2$, then $v_1 \triangleleft v_2$. Let $x$ be a maximal element of this partial order. Since $x$ has no descendants in $G$, we have that $\langle x, V \backslash (x \cup \pi(x)) | \pi(x) \rangle_G^d$, where $\pi(x)$ denotes the set of parents of $x$ in $G$. Since $G$ is a directed P-map of $\mathcal{I}$, we have that $\mathcal{I} \langle x, V \backslash (x \cup \pi(x)) | \pi(x) \rangle$. This independence statement is saturated. □

The following proposition now states that, if an independence relation includes a saturated independence statement, it must also include a saturated o-dominant statement.

**Proposition 5.3** *Let $\mathcal{I}$ be a DAG-isomorphic independence relation. Then, there exists a maximally o-dominant independence statement in $\mathcal{I}$ that is saturated.*

*Proof.* Let $G = (V, A)$ be a directed P-map of the independence relation $\mathcal{I}$. According to Lemma 5.2 there exists a saturated independence statement $\langle A, B | C \rangle$ in $\mathcal{I}$. Since $\langle A, B | C \rangle \in \mathcal{I}$, there must be a maximally o-dominant triplet $\langle X, Y | Z \rangle$ in $D_{\mathcal{I}}$ that o-dominates $\langle A, B | C \rangle$ [5, Lemma 5]. By definition of o-dominance we then have that

$$A \subseteq X \Rightarrow A \subseteq XYZ,$$
$$B \subseteq Y \Rightarrow B \subseteq XYZ,$$
$$Z \subseteq C \subseteq XYZ.$$

From $\langle A, B | C \rangle$ being saturated we further have that $ABC = V$. We thus find that $V = ABC \subseteq XYZ \subseteq V$, and hence $XYZ = V$. □

We would like to note that a DAG-isomorphic independence relation $\mathcal{I}$ may very well contain maximally o-dominant triplets that are not saturated. From the above proposition we just have that at least one maximally o-dominant triplet must be saturated.

We recall from Section 2 that any independence relation $\mathcal{I}$ can be represented by a combination of s-dominant and o-dominant triplets. We now show that if $\mathcal{I}$ is DAG-isomorphic, then *all* its maximally s-dominant triplets must be saturated. Before proving this property in Theorem 5.5 we first show that if $\mathcal{I}$ is DAG-isomorphic, then the stable part $\mathcal{S}_\mathcal{I}$ of $\mathcal{I}$ inherits the transitivity and composition properties of strong d-separation that we established in the previous section.

**Lemma 5.4** *Let $\mathcal{I}$ be a DAG-isomorphic independence relation, and $\mathcal{S}_\mathcal{I}$ its stable part. Then for any three mutually disjoint sets $X, Y, Z \subset V$ with $X, Y \neq \varnothing$, we have that*

$$\mathcal{S}_\mathcal{I}\langle X, Y|Z\rangle \Rightarrow \mathcal{S}_\mathcal{I}\langle \gamma, Y|Z\rangle \vee \mathcal{S}_\mathcal{I}\langle X, \gamma|Z\rangle,$$

*and*

$$\mathcal{S}_\mathcal{I}\langle X, Y|Z\rangle \Rightarrow \mathcal{S}_\mathcal{I}\langle X\gamma, Y|Z\rangle \vee \mathcal{S}_\mathcal{I}\langle X, Y\gamma|Z\rangle,$$

*for each $\gamma \in V \setminus XYZ$.*

*Proof.* Let $G = (V, A)$ be a directed P-map of the independence relation $\mathcal{I}$, then

$$\mathcal{S}_\mathcal{I}\langle X, Y|Z\rangle \Leftrightarrow \langle X, Y|Z\rangle_G^{S_d}.$$

The statement is now immediate from Theorem 4.2 and Proposition 4.4. □

Next we translate the transitivity and composition properties of $\mathcal{S}_\mathcal{I}$ into properties of the s-dominant statements of $\mathcal{S}_\mathcal{I}$.

**Theorem 5.5** *Let $\mathcal{I}$ be a DAG-isomorphic independence relation. Then each maximally s-dominant triplet from $\mathcal{S}_\mathcal{I}$ is saturated.*

*Proof.* From Lemma 5.2 we know that there exists a saturated independence statement $\langle X, Y|Z\rangle$ in $\mathcal{I}$. From $XYZ = V$ it follows that this statement is (trivially) stable, which implies that $\mathcal{S}_\mathcal{I} \neq \varnothing$. Since $\mathcal{S}_\mathcal{I} \neq \varnothing$, there exist maximally s-dominant triplets for $\mathcal{S}_\mathcal{I}$ [7, Lemma 3.8]. Now let $\langle A, B|C\rangle$ be a maximally s-dominant triplet of $\mathcal{S}_\mathcal{I}$ and assume that it is not saturated. Then there exists a $\gamma \in V \setminus ABC$. According to Lemma 5.4 we have that either $\langle A\gamma, B|C\rangle \in \mathcal{S}_\mathcal{I}$ or $\langle A, B\gamma|C\rangle \in \mathcal{S}_\mathcal{I}$. Since both $\langle A\gamma, B|C\rangle$ and $\langle A, B\gamma|C\rangle$ strictly s-dominate $\langle A, B|C\rangle$, this contradicts $\langle A, B|C\rangle$ being maximally s-dominant in $\mathcal{S}_\mathcal{I}$. We conclude that the assumption $ABC \neq V$ must be false, and that $\langle A, B|C\rangle$ must be saturated. □

Theorem 5.5 in essence states that a necessary condition for an independence relation $\mathcal{I}$ to be DAG-isomorphic, is that its maximally s-dominant triplets must be trivial. This condition is necessary but not sufficient. As an example we consider the independence relation

$$\mathcal{I} = \{\langle a, b|cd\rangle, \langle c, d|ab\rangle\}.$$

with $V = \{a, b, c, d\}$. The two statements in $\mathcal{I}$ are maximally s-dominant and saturated, but the relation does not have a directed P-map.

Pearl presented the following set of necessary conditions for DAG-isomorphism of an independence relation [4, Section 3.3.3]:

**C1:** $\mathcal{I}\langle X, Y|Z\rangle \Rightarrow \mathcal{I}\langle Y, X|Z\rangle$;

**C2:** $\mathcal{I}\langle X, YW|Z\rangle \Leftrightarrow \mathcal{I}\langle X, Y|Z\rangle \wedge \mathcal{I}\langle X, W|Z\rangle$;

**C3:** $\mathcal{I}\langle X, Y|ZW\rangle \wedge \mathcal{I}\langle X, W|ZY\rangle \Rightarrow \mathcal{I}\langle X, YW|Z\rangle$;

**C4:** $\mathcal{I}\langle X, YW|Z\rangle \Rightarrow \mathcal{I}\langle X, Y|WZ\rangle$;

**C5:** $\mathcal{I}\langle X, Y|Z\rangle \wedge \mathcal{I}\langle X, W|YZ\rangle \Rightarrow \mathcal{I}\langle X, YW|Z\rangle$;

**C6:** $\mathcal{I}\langle X, Y|Z\rangle \wedge \mathcal{I}\langle X, Y|Z\gamma\rangle \Rightarrow \mathcal{I}\langle X, \gamma|Z\rangle \vee \mathcal{I}\langle \gamma, Y|Z\rangle$;

**C7:** $\mathcal{I}\langle \alpha, \beta|\gamma\delta\rangle \wedge \mathcal{I}\langle \gamma, \delta|\alpha\beta\rangle \Rightarrow \mathcal{I}\langle \alpha, \beta|\gamma\rangle \vee \mathcal{I}\langle \alpha, \beta|\delta\rangle$;

for all $X, Y, W, Z \subset V$, and $\alpha, \beta, \gamma, \delta \in V$.

The conditions are termed the *symmetry* (C1), *composition/decomposition* (C2), *intersection* (C3), *weak union* (C4), *contraction* (C5), *weak transitivity* (C6), and *chordality* (C7) conditions. These conditions are satisfied by d-separation in DAG's, and hence are necessary for an independence relation to be DAG-isomorphic. Note that the conditions include the semi-graphoid axioms A1–A4.

The transitivity property of Lemma 5.4 is an extra condition additional to the list C1–C7. It thus allows to detect a larger class of independence relations that are not DAG-isomorphic. Transitivity is not implied by Pearl's conditions and it is a stronger condition than weak transitivity. The transitivity property needs to be checked only on the stable part of a relation, whereas the conditions C1–C7 must be checked on the entire independence relation.

Testing the necessary conditions for DAG-isomorphism of a semi-graphoid independence relation means checking whether all its statements satisfy the conditions C1–C7 as well as the transitivity condition stated in Lemma 5.4. In fact, only the conditions that are not the semi-graphoid axioms need to be tested, since the semi-graphoid axioms are by definition satisfied by the independence relation. The test of the remaining conditions can be performed on any representation of the relation, regardless of whether it is represented by complete enumeration of its statements, or by its set of maximally o-dominant and/or s-dominant statements. The complexity of these tests is in the order of the cube of the size of the representation. The importance of Proposition 5.3 and Theorem 5.5 lies in the fact that they provide an extra test of which the complexity is only linear in the size of the representation with dominant triplets. In a representation with maximally o-dominant triplets we need to check if there exists at least one saturated maximally o-dominant triplet. This can be done by a simple inspection of each element of this set. In a representation with maximally o-dominant and maximally s-dominant triplets we can also check if *all* maximally s-dominant triplets are sat-urated. This is also done by inspection of each element of this set.

## 6 Application to network construction

In this section we briefly discuss the practical application of strong d-separation to network construction. When building a probabilistic network, the network is preferred to be a directed P-map of the independence relation that we want to represent. During the construction phase of the network we try to determine the influences between variables, which determines the topological properties of the graph. This construction can be done either automatically through data analysis, or manually from expert interviews. In both manners of construction it is possible to detect strong conditional independence statements. A conditional independence statement can, for instance, be tested by asking a question along the lines of "do you think that knowing $Y$ is relevant to determining the value of $X$ if you know the value of $Z$?". Testing for a strong conditional independence statement would require us to ask as a second question: "and does $Y$ remain irrelevant no matter what further observations we might obtain?". If we want to obtain a directed P-map for our independence relation, then a positive answer to the latter question would imply that $X$ and $Y$ are strongly d-separated by $Z$. The transitivity property of strong d-separation then leads to a list of extra questions that can be asked: "Do you think that also $W$ is irrelevant to determining the value of $X$ or $Y$ if you know the value of $Z$?". A negative answer to such a question leads to the conclusion that there exists no directed P-map for the independence relation. A positive answer can provide an indication where in the graph the variable $W$ should be located (cf. Figure 1).

The same line of reasoning can be followed to design a series of tests that can be performed during automated model construction.

## 7 Conclusion

In this paper we introduced the concept of strong d-separation in directed graphs, and we demonstrated its relation to the concept of strong stability in semi-graphoid independence relations. We derived a set of properties for strong d-separation. These properties defined necessary conditions for a semi-graphoid independence relation to be DAG-isomorphic. These properties can be implemented as a test procedure in model construction. We also showed that the combined properties lead to a test that can be performed on a representation of a semi-graphoid independence relation by dominant triplets. The complexity of this test is linear in the size of the representation of the relation.

Strong d-separation is a translation of stable independence onto a directed P-map. We plan to investigate if more properties of the topology of the graph can be derived that are due to stability. We also foresee to study the influence of these properties on the computational aspects of inference. A possible direction may be that they lead to special properties of the chordal graph, which may have an impact on clustering in the junction tree algorithm [2].

## 8 Acknowledgement

This research was (partly) supported by the Netherlands Organisation for Scientific Research (NWO). We would like to thank the anonymous reviewer who provided helpful suggestions with respect to the proof of Theorem 4.2.